\documentclass[sigchi-a, authorversion]{acmart}
\usepackage{booktabs} 
\usepackage{ccicons}  

\setcopyright{acmcopyright}

\acmDOI{}

\acmISBN{}

\acmConference[CHI 2019, Glasgow]{Where is the Human? Bridging the Gap Between AI and HCI Workshop}{May 2019}{Glasgow}
\acmYear{2019}
\copyrightyear{2019}

\acmPrice{15.00}

\begin{document}
\title{Transparency in Maintenance of Recruitment Chatbots}

\author{Kit Kuksenok}
\affiliation{%
  \institution{jobpal}
  \city{Berlin}
  \country{Germany} }
\email{kit@jobpal.ai}

\author{Nina Pra{\ss}}
\affiliation{%
  \institution{jobpal}
  \city{Berlin}
  \country{Germany} }
\email{nina@jobpal.ai}

\renewcommand{\shortauthors}{F. Author et al.}

%
%

\ccsdesc[500]{Human-centered computing~Natural language interfaces}
\ccsdesc[300]{Human-centered computing~HCI design and evaluation methods}
\ccsdesc[300]{Human-centered computing~Empirical studies in HCI}

\begin{abstract}
We report on experiences with implementing conversational agents in the recruitment domain based on a machine learning (ML) system.  Recruitment chatbots mediate communication between job-seekers and recruiters by exposing ML data to recruiter teams. Errors are difficult to understand, communicate, and resolve because they may span and combine UX, ML, and software issues. In an effort to improve organizational and technical transparency, we came to rely on a key contact role. Though effective for design and development, the centralization of this role poses challenges for transparency in sustained maintenance of this kind of ML-based mediating system.


\end{abstract}

\keywords{Conversational agent; Conversational UI; implications for design; mixed-initiative system.}

\maketitle

\section{Introduction}

\begin{marginfigure}
    \includegraphics[width=\marginparwidth]{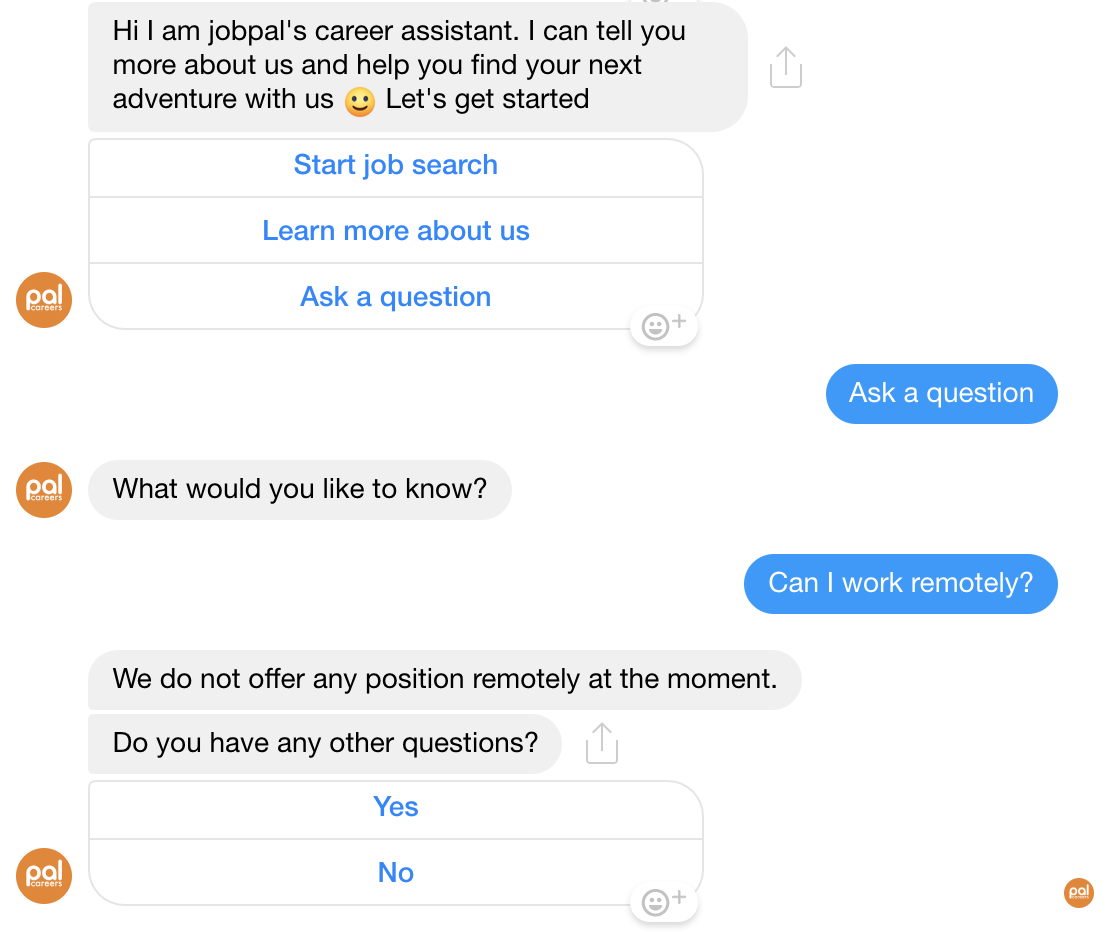}
    \caption{In the FAQ interaction, a question from a job-seeker receives an answer immediately, based on an ML model built on all prior questions and answers. If the confidence level is too low, the question will be escalated to the recruiters. \\ }
    \label{fig:sampledia}
    
    \includegraphics[width=\marginparwidth]{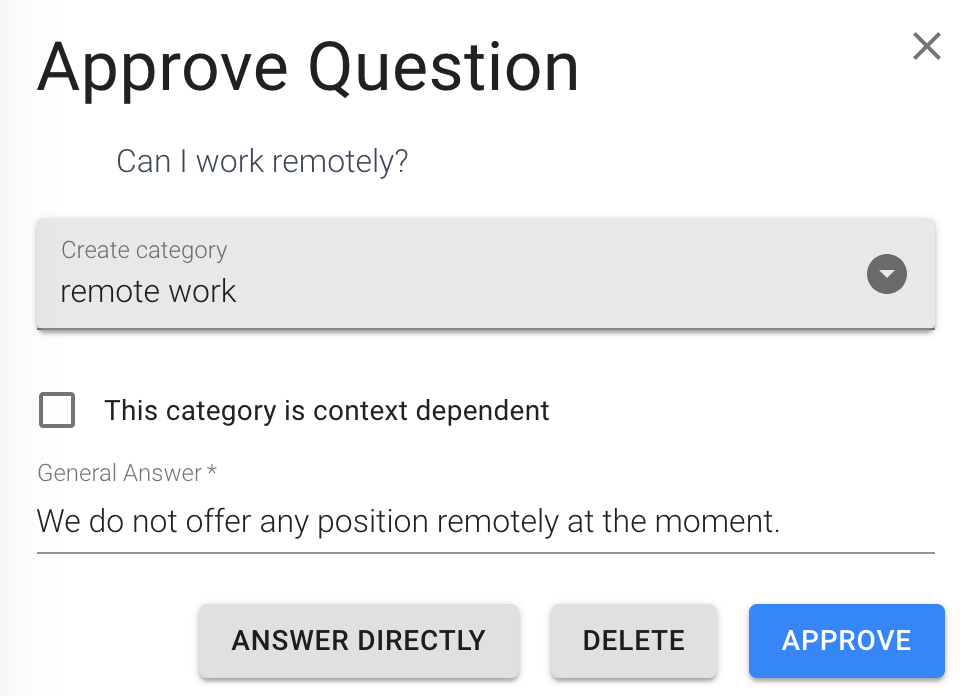}
    \caption{Even if the chatbot responds, the recruiters can use a dashboard  to review the the answer.}
    \label{fig:dbcontrol}
\end{marginfigure}

A smart conversational agent connects an end-user and a domain expect through a machine learning (ML) system. We focus on the recruitment domain, but such smart conversational agents are increasingly  explored across business domains (e.g., ~\cite{jain2018evaluating}, ~\cite{kuligowska2015commercial}), where our experiences may also apply. Behind a dialog like the one in Figure~\ref{fig:sampledia}, a small team of recruiters can maintain a dataset of frequently-asked questions (FAQs) and corresponding answers, by reviewing all responses provided by the chatbot (Fig.~\ref{fig:dbcontrol}). In the perfect world, the technology seamlessly helps to support both parties. Interacting with an automated mediator, a job-seeker may feel more free to ask questions that they would not ask a recruiter directly, such as questions about salary, parental leave, and workplace diversity. Semi-automated FAQs allow the staff to focus on more complex queries.

The promise of this kind of  technology may motivate  initial buy-in, but sustained use requires transparency. Existing chatbot tools\textemdash both high-level platforms \cite{canonico2018comparison} and lower-level technical strategies \cite{abdul2015survey}\textemdash do not actively support estimating ML errors. Chatbot guidelines leave error handling to the designer: ``transparency'' is included as an important topic for error handling in particular, meaning to ``be honest and transparent when explaining why something doesn't work'' \cite{dialogflowerrors}. In practice, why something does not work, and under what conditions, can puzzle the designer or developer, not just the end-user.

The information needed for transparency, as defined above, is unavailable for technical and organizational reasons. Developers may tend to overlook the data, in favor of focusing on algorithmic improvements and errors in building ML systems \cite{patel2008examining}. Errors combine software bugs, data quality problems, conceptual misunderstanding, and/or conversational flow design issues. We identified the information needs and information tasks of the key role who acts as a liaison between the developer team and the recruitment staff on a particular client project  (Fig.~\ref{fig:chalstrat}).
The further challenge is to understand how to more effectively organize the work and distribute it across other, existing  roles within such projects, beyond initial design and development.

\section{Key contact role: information needs and tasks}

To construct  the  model shown in Figure 3, we drew from experiences with two illustrative projects. Both were implemented at companies with over $40K$ employees, with a commensurate hiring rate. The training set of the FAQ portion of each project contains between $1.5K$ and $5K$ training examples across between $100$ and $200$ distinct classes. Both datasets consistently perform at around $70\%$ accuracy on this multi-class classification problem, using logistic regression and language-specific models.  The datasets were English and German, respectively. Over the last three months of operation, one of the chatbots saw about \textbf{$2000$} active users, of which about \textbf{$250$} (around $ 12\% $) asked questions. The other saw about \textbf{$65$} weekly active users, of which \textbf{$15$} (around $23\% $) asked questions. These projects represent the opposite ends of the traffic spectrum of our projects overall. In both cases, a typically small number (2-4) of recruiters were ultimately responsible for maintaining the ML dataset.

\begin{figure}
    
    \includegraphics[width=230pt]{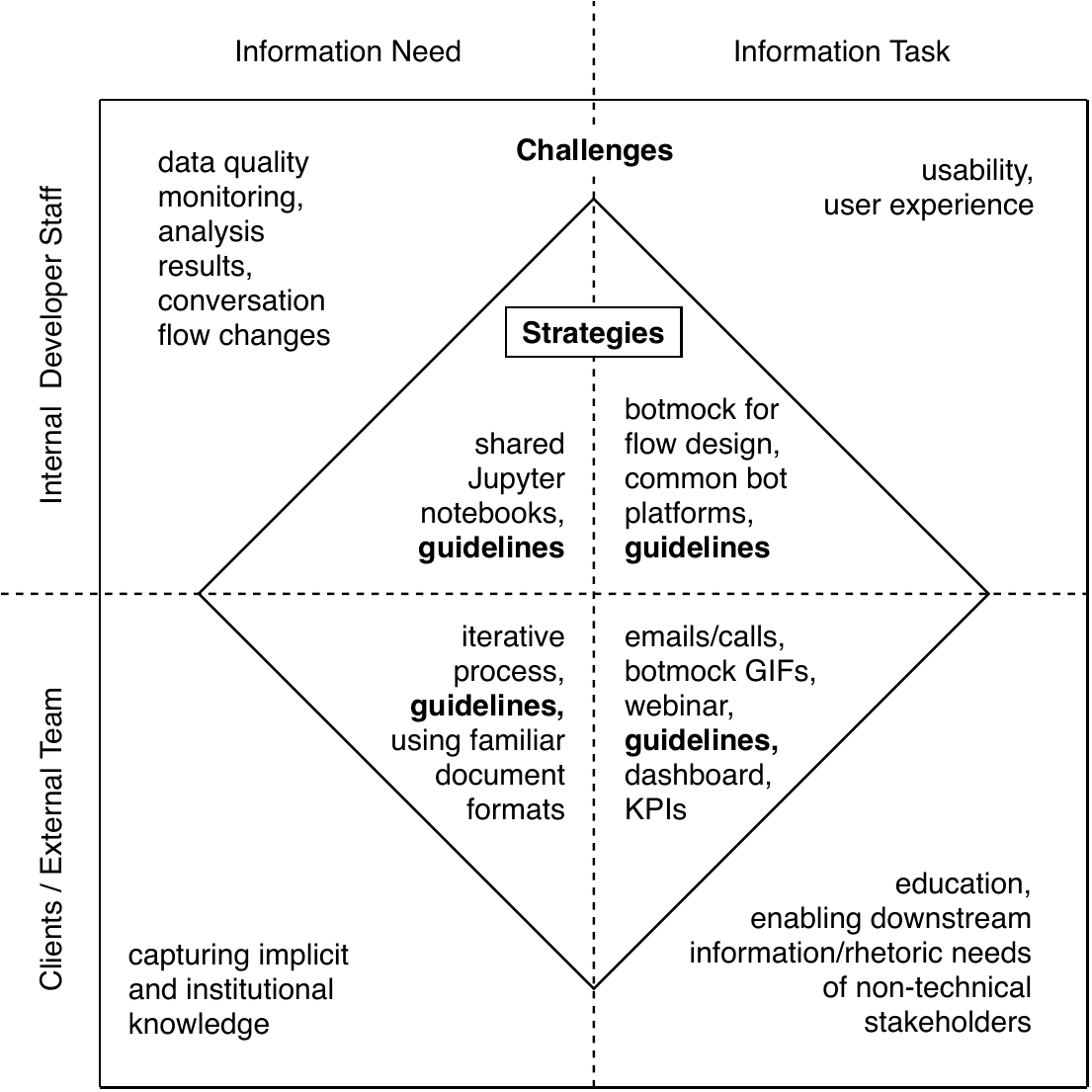}
    \caption{The key contact role has internal and external information needs and tasks. Strategies include shared guidelines co-created by different stakeholders.}
    ~\label{fig:chalstrat}
\end{figure}

During initial design and development, shared documents are used differently by different stakeholders. To gather initial FAQ data, the PM and the client iterate using spreadsheet documents first, starting with familiar document formats to focus on the data quality concepts before onboarding with specialized tools. To define the overall conversational flow, the PMs use a tool that creates a flow overview, as well as an animation of going through the conversation flow. The flow is useful when the PM develops the flow with the client, and when the PM creates a ticket for a developer; as well as during hand-off in the  client organization. The animation is useful for internal rhetorical needs on the client's side for sustaining the project.

Deployment of the chatbot brings real data from real job-seekers. The recruiters then use the internal dashboard to train and maintain the ML dataset. This may reveal dramatic differences between anticipated and actual use of  the chatbot, and results in potentially significant revisions. The PMs, who play the  key contact role (Fig.~\ref{fig:chalstrat}), use a variety of analytic tools to understand, communicate, and  resolve problems. For example, to address data quality errors, PMs run analysis scripts in Python (in the  Jupyter Notebook environment) and suggest areas for improvement to the client. Their synthesizing work forms a bridge between the development team who maintains the notebooks, and the clients who are invested in practical solution to specific problems.

\section{Future Work}

\begin{sidebar}
    Which questions would you like not to answer anymore? \\[1\baselineskip]
    These frequently asked questions are likely to be already in:\\
    ...your current FAQ list, internally and on your website\\
    ...e-mails from jobseekers, previous chat solutions, phone\\
    ...social media channels (e.g., Facebook Career Page) \\[1\baselineskip]
    It helps to involve the department(s) which already deal\\with answering these questions on a daily basis.
    \caption{Suggestions on capturing existing FAQs for initial data, taken verbatim from a guideline document, created with the developer team and shared with the client in a webinar.}
\end{sidebar}
The multiplicity of uses and affordances by shared  documents in multi-stakeholder settings has been the subject of prior HCI work. In our experience, these familiar issues are exacerbated in the context of an ML system,  when the non-developer stakeholders (i.e., recruiters) are responsible for data quality maintenance. The extensive responsibilities of the key contact role beg the question: how can this work be more effectively de-centralized, and sustained? 

In the last quarter, jobpal chatbots answered $70\%$ of incoming questions without human escalation; then, of the guesses made, about $60\%$ are approved (as in Figure 2) without correction by recruiters. This  means that  the  current level of organizational and technical transparency involves a great deal of passive review and agreement with  machine suggestions. Practical, sustained transparency requires an actionable and engaging UI to  the  ML system. In one major direction of future work, we are building on work from CHI, InfoVis, and VAST communities to move analytic and strategies components into the client-facing UI. This includes adapting research on ML understandability to the textual, natural language processing (NLP) domain.

\begin{acks}
We thank Klaudia Niedba\l{}owska and Flora Wie{\ss}ner, who were interviewed as part of this case study.
  
\end{acks}

\bibliography{sample-bibliography-sigchi-a}
\bibliographystyle{ACM-Reference-Format}

\end{document}